\documentclass{sig-alternate-05-2015}

\usepackage{hyperref}
\usepackage{textcomp}

\usepackage{balance}

\begin{document}

\setcopyright{acmcopyright}

\doi{10.475/123_4}

\isbn{123-4567-24-567/08/06}

\conferenceinfo{KDD '16}{August 13--17, 2016, San Francisco, CA, USA}

\acmPrice{\$15.00}

%
\conferenceinfo{KDD '16}{August 13--17, 2016, San Francisco, CA, USA}

\title{Applying Deep Learning to Basketball Trajectories}

%
%
%
%
%

\numberofauthors{2} 
%
\author{
%
%
\alignauthor
Rajiv C. Shah\\
       \affaddr{University of Illinois at Chicago}\\
       \affaddr{Department of Communication}\\
       \email{rshah@pobox.com}
\alignauthor
Rob Romijnders\\
       \affaddr{Eindhoven University of Technology}\\
       \email{romijndersrob@gmail.com}
}

\maketitle
\begin{abstract}
One of the emerging trends for sports analytics is the growing use of player and ball tracking data. A parallel development is deep learning predictive approaches that use vast quantities of data with less reliance on feature engineering. This paper applies recurrent neural networks in the form of sequence modeling to predict whether a three-point shot is successful. The models are capable of learning the trajectory of a basketball without any knowledge of physics. For comparison, a baseline static machine learning model with a full set of features, such as angle and velocity, in addition to the positional data is also tested. Using a dataset of over 20,000 three pointers from NBA SportVu data, the models based simply on sequential positional data outperform a static feature rich machine learning model in predicting whether a three-point shot is successful. This suggests deep learning models may offer an improvement to traditional feature based machine learning methods for tracking data.
\end{abstract}

%
%

\ccsdesc[500]{Computing methodologies~Neural networks}

%
%

%
%
\printccsdesc


\keywords{Deep learning; recurrent neural networks, SportVu, basketball, tracking, trajectories,}

\section{Introduction}
This paper classifies three point shots based solely on tracking data. This is done by using a recurrent neural network (RNN) that learns sequences of movements. RNNs are a class of dynamic models used to predict and generate sequences in domains such as text \cite{Martens2011}, music \cite{Liu2014}, and motion data \cite{Alahi2016}. 
\par
The inspiration for applying RNNs to ball tracking data stems from the work of Graves who uses RNNs to develop predictions on handwriting \cite{Graves2013}. Graves used XY sequential data taken from handwriting on a \textit{smart whiteboard}. He proceeded to train a RNN network on the XY data without any preprocessing. The model was then able to not only predict the next letter or word, but even generate sequences based on different initial starting points. Graves offers an online handwriting demo that allows anyone to better understand the significance and potential of the work\footnote{See http://www.cs.toronto.edu/~graves/handwriting.html}.
\par
Imagine applying this generative model to sports tracking data to predict player/ball movement that would allow generating dynamic sequences that reflect the tendencies of a specific player. For example, it could be possible to create fictional scenarios, but have ball and player movement in the style of a player, e.g, penetration drives based on the style of Jeremy Lin.
\par
An attempt by Wang to use RNNs on tracking data was unsuccessful \cite{Wang2016}. While he was able to use RNNs on images, he notes that using the difficultly of using the XY positional data. He suggests that pictorial representation is a better method for analyzing plays. Based on Wang\textquotesingle s warnings, the authors approached the idea of RNN cautiously by assessing the ability of RNNs to learn sequence data in two ways\cite{shiny}.

\begin{figure}[h]
    \centering
    \includegraphics[width=0.8\linewidth]{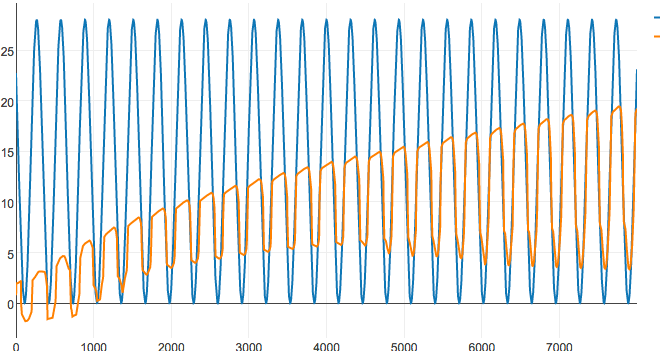}
    \caption{RNN learning a sine wave}
    \label{intro}
\end{figure}

As figure \ref{intro} shows, over time the RNN learns and anticipates the shape of the wave. After enough training cycles, the network learns the shape and can anticipate well.

\par

The second way is to learn simple addition. A simple RNN model learned to add between 5 to 15 single digit numbers \cite{RNN1st}. For example, using a 2 layer LSTM network with 100 hidden units, a batch of 50 training examples, and 5000 epochs, the RNN is able to sum:
\[8+6+4+4+0+9+1+1+7+3+9+2+8 \text{ as } 66.215\]
This isn't too far from the actual answer of 62. Further training can improve the performance. 
\par
Based on those positive results, we focus on predicting ball movement on three-point shots for several reasons. First, ball movement has a much higher velocity than players and therefore is more difficult to model. Second, the trajectory of ball movement is non-linear, which makes RNN a better fit than traditional linear models. Third, little attention has been given to ball movement, while there is a large scholarship around extracting player movement from tracking data. The rest of this article discusses the approach, experimental results, and the implications of our classifier for three-point attempts based on tracking data.

\section{Approach}
\subsection{Data preparation}
The data used in this study stems from publicly available SportVu data. SportVu is an optical tracking system installed by the National Basketball Association (NBA) in all 30 courts to collect real-time data. The tracking system records the spatial position of the ball and players on the court 25 times a second during a game.
\par
\begin{figure}[h]
    \includegraphics[width = \linewidth]{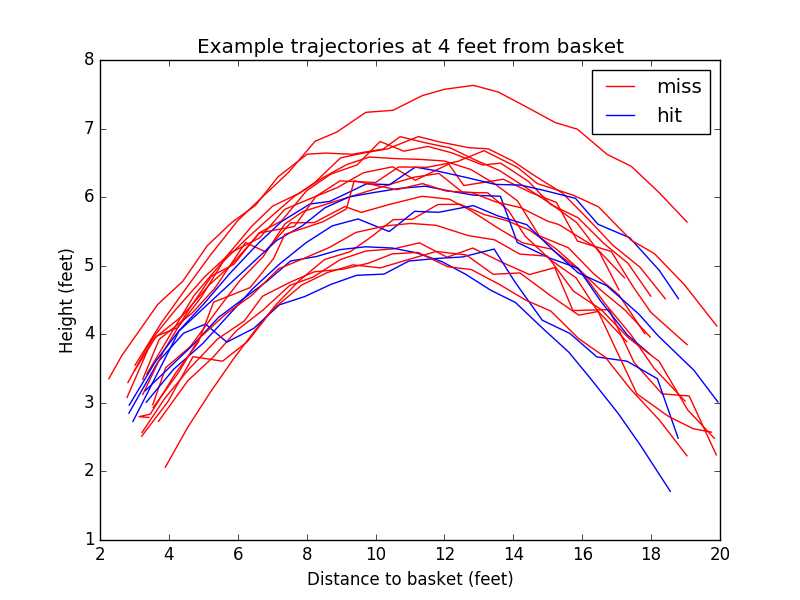}
    \caption{Examples of basketball trajectories in the dataset\textit{Trajectory until the basketball i within 4 feet of the basket}}
    \label{2d}
\end{figure}

This study focused  on three point plays as defined by ball movement over at or greater than 8 feet in height and over a range of 22 feet in the SportVu tracking data. This data was joined with the play by play data from the NBA, which indicates when a three-point shot is taken and whether it is successful. Only shots that are in both datasets are kept. Figure \ref{2d} shows examples of trajectories in our dataset. The data in the figure only shows the trajectory prior to a distance of four feet from the basket. The height refers to the distance above the basketball rim. Additionally, the X and Y coordinate are combined into a single distance.

\par
The dataset for this study consists of over 20,000 three point shot attempts from 631 games. The data was taken from the NBA.com site in the beginning of the 2015-2016 season. The dataset does not contain every three-point attempt in those 631 games. The incompleteness of the publicly available SportVu data and only keeping verifiable shots with both play by play limits the size of our dataset. The percentage of made shots in the dataset is 35.7\%, which compares favorably to the 35\% season average for 2015-2016 regular season \cite{BR}.
\par
The first dataset consists of only the X, Y, Z, and game clock variables representing the location of the ball in three dimensions over time. X refers to the length of the court, Y is the width of the court, and Z is the height of the ball. A second dataset is created with additional variables based on the physics of ball trajectories. The belief was that these variables would add more information over just the location data for machine learning models. Specifically, the added variables included the difference in movement over each time period for each dimension. Three other variables included: the distance to the center point of the rim, the difference over time for this distance, and the angle of the ball with respect to the rim. 
\par
Data for both datasets is centered.  Additionally, the data is split into a train/test datasets using a split of 80/20.  
\subsection{Recurrent Neural Network}
This paper will forgo the mathematical formulas associated with RNNs. For those seeking a fuller treatment, we refer you to the work of Schmidhuber \cite{Schmidhuber2015}.
\par
In this study, we use a popular variant of RNN with long short-term memory (LSTM) units. The network architecture relies on a two layered LSTM using peephole connections. The input to the LSTM is the XYZ data and the game clock. At each time step, the RNN predicts both the probability of a successful shot and parameters for the mixture density network (MDN). The probability comes from a softmax layer and is trained based on cross entropy error. The MDN consists of three mixtures of tri-variate Gaussians and is trained via cross entropy.
\par
An Adam optimizer was used along with dropout in the models. The results for the classification model use the area under the receiver-operating characteristic curve known as AUC. AUC can range from 0.5 (pure chance) to 1.0 (ideal classification).
\section{Experimental results}
The goal is to predict whether a shot is a make or miss between two to eight feet from the basket. As the ball is further from the basket, there is more uncertainty for the models to consider. This section provides results on a baseline model using non-sequential data and a RNN using sequential data.
\subsection{Baseline models}
To assess the value of a sequencing model, the first step was setting a baseline using traditional techniques. Using the last point (closest to the basket), we built classifiers using a generalized linear model and gradient boosted machines (GBM). These classifiers provide insight into how valuable the last data point is as well as possible interactions between variables. The parameters for the classifiers relied on the default values and were not optimized. The goal was a rough approximation of how a non-sequential model would perform.

\begin{table}[h]
\begin{center}
    \begin{tabular}{| p{2cm} | l | l |}
    \hline
       & \textbf{GLM} & \textbf{GBM} \\ \hline
   AUC & 0.53 & 0.80 \\ \hline
    \end{tabular}
    \caption{Baseline XYZ models at 1 foot from basket}
    \label{table1}
    \end{center}
\end{table}

\par
The first set of models only used the three variables that indicated the position of the ball at one foot above the basket: X, Y, Z. A logistic regression using Elastic Net with an alpha of 0.5 resulted in an AUC of 0.53. A gradient boosted trees model with 50 trees resulted in an AUC of 0.80 as shown in Table 1. As the later models will highlight, this performance is much worse than other approaches.
\par
To improve the performance of these models, additional variables were added that experience suggests should improve a trajectory model. These variables included: X, Y, Z coordinates of the ball, the distance to the center of the basket, the difference between the last two points for X, Y, Z, distance to center, and the angle of the ball with respect to the basket. The results in Table \ref{table2} are for an Elastic Net with an alpha of 0.5 and a gradient boosted trees model with 50 trees.
\begin{table}[h]
\begin{center}
    \begin{tabular}{| p{2cm} | l | l |}
    \hline
    \textbf{Distance to basket} & \textbf{GLM} & \textbf{GBM} \\ \hline
    2 feet & 0.875 & 0.942 \\ \hline
3 feet & 0.807 & 0.902 \\ \hline
4 feet & 0.721 & 0.848 \\ \hline
5 feet & 0.659 & 0.796 \\ \hline
6 feet & 0.604 & 0.746 \\ \hline
7 feet & 0.583 & 0.742 \\ \hline
8 feet & 0.558 & 0.719 \\ \hline
    \end{tabular}
    \caption{AUC for Baseline models on full feature dataset }
    \label{table2}
    \end{center}
\end{table}

\par
It is interesting to note that the tree based models perform much better. This indicates that the data contains non-linearities. 
\subsection{RNN with only positional dataset}
The RNNs are fed a sequence length of 12, which represents about a half a second of time. The inputs consists of the three positional dimensions and game clock. The network is then trained and scored on the validation set. The results are shown in Table 5. The network architecture uses 2 layers with a LSTM of 64 hidden units, Adam optimizer with a learning rate of 0.005, dropout of 0.6, and a batch size of 64. 
\begin{table}[h]
\begin{center}
    \begin{tabular}{| p{2cm} | l | l |}
    \hline
    \textbf{Distance to basket} &\textbf{RNN} \\ \hline
    2 feet & 0.93\\ \hline
3 feet & 0.913 \\ \hline
4 feet & 0.906 \\ \hline
5 feet & 0.880 \\ \hline
6 feet & 0.873 \\ \hline
7 feet & 0.841 \\ \hline
8 feet & 0.843 \\ \hline
    \end{tabular}
    \caption{AUC for RNN models on the positional dataset }
    \label{table3}
    \end{center}
\end{table}

\par
The RNN is able to improve from baseline models at all distances except 2 feet.  The AUC values for distances from 6 to 8 feet are impressive. They are considerably better than the GBM and show that the RNN was able to learn and classify basketball trajectories.  

\section{Discussion}

There are a number of interesting issues arising from the experimental results. The first are the implications of the results, particularly the performance of the RNN on the positional dataset. This result speaks directly to the role of feature engineering when using deep learning. The second issue focuses on possible ways to improve the RNN performance. The final issue considers the limitations on performance due to the SportVu data. 
\par
The RNN is able to produce the highest classification scores. Only using positional data, these models outperform the feature engineered GBM models. This suggests the sequential RNN models are capable of learning nonlinear behavior. While this application may be considered simplistic, this work illustrates just how well RNNs can learn sequential behavior.
\par
The second issue concerns methods for improving performance for RNNs. The first step would be a more comprehensive search for better performing hyper-parameters. The models in this paper are not fully optimized. Another approach is increasing the training set size. In this study, we found that at 4 feet we could reach an AUC of 0.870 with just half the training data versus 0.906 with the full training set. It is readily apparent that ball trajectories are a much simpler problem than other predictive models, such as play prediction. This suggests for more complex prediction tasks larger datasets will be more beneficial. Consider the training sets used in other RNN applications. For example, in Graves' work on handwriting used XY data to predict the next letter or sample. The IAM online handwriting dataset consists of over 85,000 words, each of words is broken up into line strokes with xy and time. The resulting dataset for just the line strokes is about 500 MB. In the character level RNN work, the training text includes over 5 million characters \cite{Karpathy2016}. In contrast, we are looking at tens of thousands of shots which total about 40 MB. One method to ameliorate the paucity of the training dataset is augmenting by reworking the existing data. An example of applying this can be found in a recent winner of a Kaggle competition, who noted the "canonical examples are found in image classification tasks where images are cropped and perturbed to improve the generalization capabilities of the classifier." \cite{pupa} Their team was able to apply similar techniques to their sequential data to augment it.
\par
\begin{figure}
    \centering
    \includegraphics[width=0.8\linewidth]{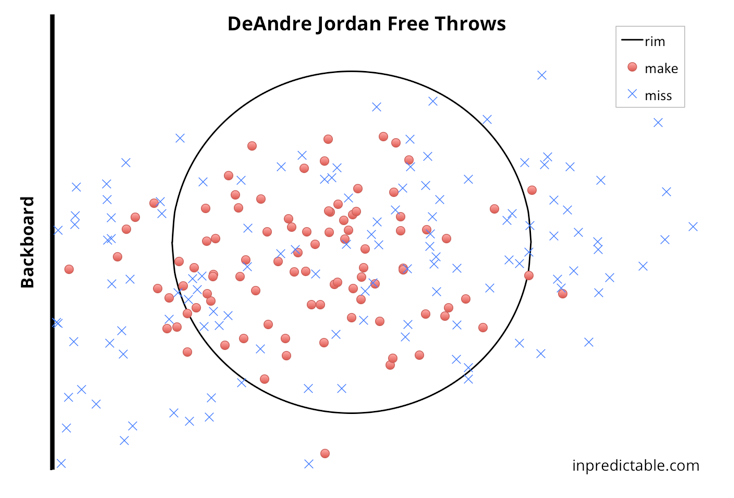}
    \caption{Physics based free throw shot predictions }
    \label{pic_disc}
\end{figure}

The last issue to consider is the limits of performance due to the SportVu data. The SportVu data is based on optical tracking at a rate of 25 times a second. Close inspection of most ball trajectories shows they are not entirely smooth, but can involve "dips" or noise. A blog post by Mike Beuoy provides insight into the performance of ball trajectories using SportVu data \cite{sharc}. His work analyzed over 30,000 free throw shots with a physics based model that looked at four main forces on a ball: gravity, buoyancy, drag, and the magnus effect. He predicts the location of the ball as shown in Figure \ref{pic_disc}
\par

There are several misses on this chart that appear within the area of the rim. Beouy suggest the following explanation, "While this could be due to a shallow approach angle, the more likely, and less interesting explanation is that the SportVU data is simply messy and imprecise (to say nothing of my own imperfect methodologies for deciphering said data)." The distance and differing angles of a three point shot compared to the free throw shots suggest it could even have a higher error factor, limiting the ability of a model to make ball tracking predictions.

\balance

\section{Conclusion}
This paper develops neural network models for classifying the trajectory of the ball. A RNN network had the best performance over traditional static approaches. The RNN is able to achieve an AUC of 0.843 when predicting a make or miss using half a second of data with the ball 8 feet away from the basket. This outperforms the traditional approaches which had an AUC of 0.558 and 0.719 for a general linear model and a gradient boosted machines, respectively.
\par
This paper focuses on a simpler problem by solely focusing on three-point trajectories. However, it is not readily apparent given the high ball velocity and noisy nature of the motion data, whether a sequential classifier would add value. The results here clearly indicate RNNs offer value.
\par
This paper stands in the vanguard for applying RNNs to motion tracking data. The results here suggest RNNs have the ability to offer an improved understanding of sequential data. Future work will likely study other motion tracking tasks, such as play classification or even the individual style of a player. In other contexts, such as handwriting, a RNN can learn the style of a person. In the same way, we are hopeful that RNNs can be used to learn the style of an individual basketball player. 

%
\bibliographystyle{abbrv}
\bibliography{library}  
%
%

\section{Code/Data}
A short summary of this paper is available at 
\href{http://tinyurl.com/traj-rnn}{tinyurl.com/traj-rnn}. 
You can also download the data and the model used for this paper on
\href{https://github.com/RobRomijnders/RNN_basketball}{github}. 

\end{document}